\documentclass[10pt,twocolumn,letterpaper]{article}

\usepackage{cvpr}              

\usepackage[dvipsnames]{xcolor}

\usepackage{indentfirst}
\definecolor{cvprblue}{rgb}{0.21,0.49,0.74}
\usepackage[pagebackref,breaklinks,colorlinks,citecolor=cvprblue]{hyperref}
\usepackage{multirow}

\def\paperID{4685} 
\def\confName{CVPR}
\def\confYear{2024}

\title{Learning Intra-view and Cross-view Geometric Knowledge for Stereo Matching}

\usepackage{authblk}

\author[1]{Rui Gong\thanks{The work was done during an internship at $\text{I}^2\text{R}$, A*STAR.}}
\author[2]{Weide Liu}
\author[2]{Zaiwang Gu}
\author[2]{Xulei Yang}
\author[2]{Jun Cheng\thanks{cheng\_jun@i2r.a-star.edu.sg}}
\affil[1]{School of Electrical and Electronic Engineering, Nanyang Technological University}
\affil[2]{Institute for Infocomm Research, A*STAR}

\begin{document}
\maketitle
\begin{abstract}

Geometric knowledge has been shown to be beneficial for the stereo matching task. However, prior attempts to integrate geometric insights into stereo matching algorithms have largely focused on geometric knowledge from single images while crucial cross-view factors such as occlusion and matching uniqueness have been overlooked. To address this gap, we propose a novel Intra-view and Cross-view Geometric knowledge learning Network (ICGNet), specifically crafted to assimilate both intra-view and cross-view geometric knowledge. ICGNet harnesses the power of interest points to serve as a channel for intra-view geometric understanding. Simultaneously, it employs the correspondences among these points to capture cross-view geometric relationships. This dual incorporation empowers the proposed ICGNet to leverage both intra-view and cross-view geometric knowledge in its learning process, substantially improving its ability to estimate disparities. Our extensive experiments demonstrate the superiority of the ICGNet over contemporary leading models. The codes will be available at \url{https://github.com/DFSDDDDD1199/ICGNet}.
\end{abstract}    
\section{Introduction}
\label{sec:intro}

Stereo matching is the task of estimating a disparity map from a pair of rectified images. It stands as one of the cornerstone challenges in a range of computer vision tasks, including augmented reality, autonomous driving, and robotics \cite{usage}. With the rapid development of deep learning techniques and network structures in computer vision scenarios, deep stereo matching networks have achieved great success in stereo matching. 

\begin{figure}[h]
    \centering
    \includegraphics[width=0.48\textwidth]{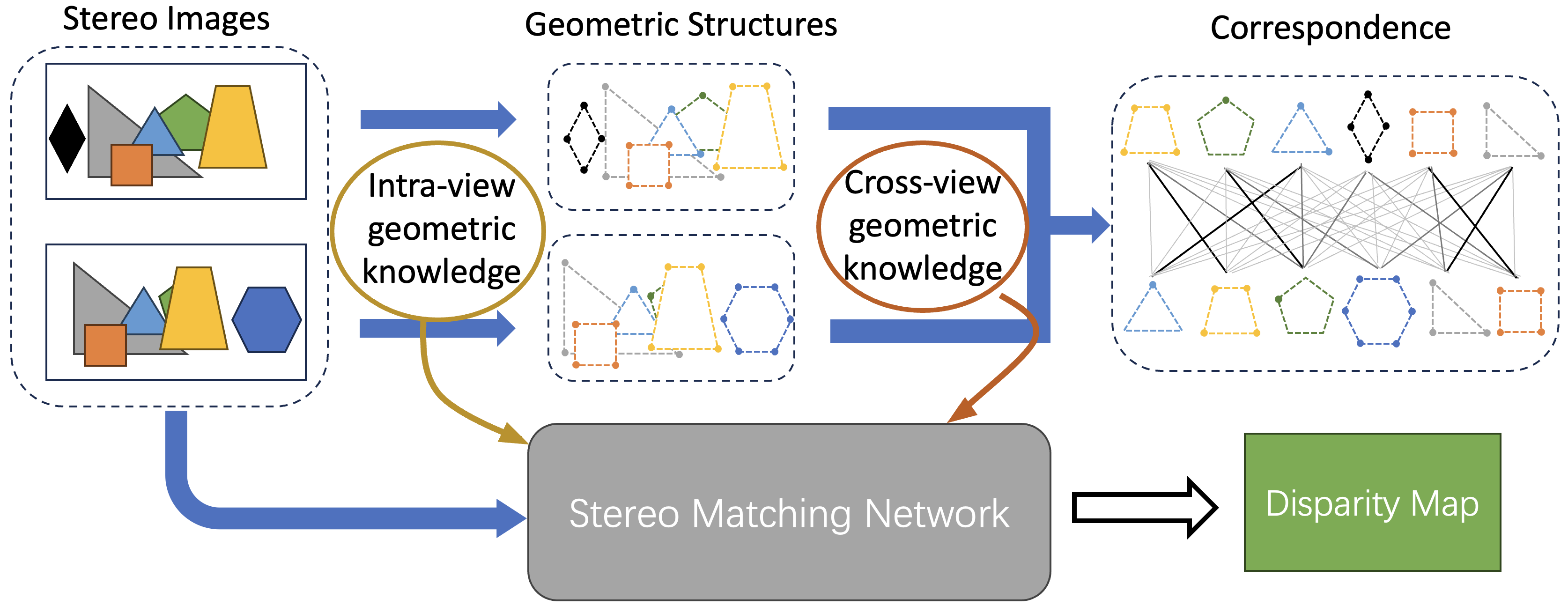}
    \caption{An illustration of our overall framework using  synthetic shapes. We leverage intra-view geometric knowledge which is the knowledge of extracting geometric structures, and cross-view geometric knowledge which is the knowledge of the correspondences of these structures, to aid the stereo matching task. Note that the dotted lines are used for illustration and  are not used in the method.}
    \label{fig:opening}
\end{figure}

 Geometric knowledge plays a crucial role in the overall performance of stereo matching networks \cite{igevstereo,edgestereo,zhang2017normal,kusupati2020normal} due to the existence of repeated patterns, textureless areas, and reflective surfaces, which make simple texture-based matching methods ineffective. Consequently, several methods were proposed to introduce geometric knowledge into stereo matching networks through kinds of geometric structures. EdgeStereo \cite{edgestereo} proposed a learning network that incorporates edges and edge-disparity relations into the stereo matching network. Normal assisted stereo matching \cite{zhang2017normal, kusupati2020normal} leverages surface normal as geometric constraints to reduce the matching ambiguity for textureless regions. However, the geometric knowledge employed in these studies is mainly intra-view, lacking the cross-view geometric knowledge such as matching uniqueness and occlusion. 

To address the absence of cross-view geometric knowledge, we have identified that the local feature matching task \cite{lowe2004distinctive,superpoint,sarlin2020superglue} can serve as a suitable source of cross-view geometric knowledge for stereo matching. Local feature matching involves the identification of correspondences between two images, typically achieved by initially detecting sparse, geometrically meaningful interest points within each view and then extracting descriptors for these interest points encoded with their visual attributes \cite{superpoint}. Subsequently, cross-view matching is carried out by comparing the interest points between the two images \cite{sarlin2020superglue}. This matching process solves a partial assignment problem between the points which takes matching uniqueness and occlusion into account. Moreover, we also find that the interest points from each image also provide intra-view knowledge of the image. Building upon these insights, we introduce our ICGNet framework to take both the intra-view and cross-view geometric knowledge contained in local feature matching models into the stereo matching network. We leverage a pre-trained local feature matching pipeline, comprising an interest point detector and a point matcher. We propose an intra-view constraint that the features we extract from stereo matching backbone preserve the same interest point information as the pre-trained interest point detector to gain intra-view geometric knowledge. Additionally, we propose a cross-view constraint that these features are compelled to serve as descriptors for the interest points, facilitating the matching process between these points. The matching outcome is then constrained to align with both the ground-truth matching labels provided by ground-truth disparity maps and the matching results obtained from the pre-trained interest point matcher, in order to gain cross-view geometric knowledge. Compared to EdgeStereo \cite{edgestereo} and normal assisted stereo matching \cite{kusupati2020normal,zhang2017normal}, our method not only additionally introduces cross-view geometric knowledge but also keeps zero overhead when inference.

Our experiments show that our method effectively improves the performance across both the seen and unseen domains. The contributions are summarized as follows:
\begin{enumerate}
    \item We propose to learn intra-view geometric knowledge for stereo matching networks from an interest point detector by introducing a novel intra-view geometric constraint computed from the interest point detection outcomes.
    \item We propose to learn cross-view geometric knowledge for stereo matching networks from an interest point matcher by introducing a novel cross-view geometric constraint. We compute both soft cross-view and hard cross-view geometric losses using pretrained interest point matcher and disparity ground truth. 
    \item We conduct a series of comprehensive experiments that consistently demonstrate the proposed framework's ability to enhance the overall performance of state-of-the-art stereo matching networks. Our ICGNet achieves state-of-the-art performance on the SceneFlow dataset and ranks \textbf{$1^{st}$} on the KITTI 2015 stereo matching benchmark among published peer-reviewed methods at the time of submission, and improves cross-domain generalization of baseline methods on KITTI 2012, KITTI 2015, and Middlebury 2014 datasets. 
\end{enumerate}
\section{Related Works}
\label{sec:related}
\subsection{Deep Stereo Matching}
Deep-learning-based deep stereo matching techniques have gained widespread adoption and demonstrated promising outcomes \cite{stereo1,stereo2,stereo3,stereo4,stereo5,stereo6,stereo7,stereo8,sttr, unifying,cstr,cspn}. The initial instance of such an approach is DispNetC \cite{sceneflow}, which presents an end-to-end trainable stereo matching framework. This framework employs the dot product of feature maps from left and right images as correspondence to build a cost volume of three dimensions. However, the drawback of this computational-friendly methodology is its inability to capture sufficient information to achieve satisfactory results. In the pursuit of improved performance, subsequent methods like GC-Net \cite{gcnet} have emerged. Numerous works \cite{psmnet,ganet} have followed this trend and incorporated 3D hourglass convolutions to aggregate a 4D cost volume. This volume is created by concatenating features from left and right images. Unfortunately, this approach demands substantial memory usage and computational complexity. A notable enhancement arrives with GwcNet \cite{gwcnet}, which introduces the concept of group-wise correlation. This innovation enables the construction of a more compact cost volume. Recently, iterative optimization-based methods \cite{crestereo,raftstereo,parameterized} have achieved splendid performance. Inspired by RAFT \cite{raft}, RAFT-Stereo \cite{raftstereo} iteratively updates the disparity using all-pairs correlation between features extracted. IGEV-Stereo \cite{igevstereo} further boosts the performance of stereo matching network by combining cost-volume modules with iterative updating modules. Uncertainty-based methods \cite{uncertainty1,uncertainty2} also gained attention these years. ELFNet \cite{elfnet} leverage uncertainty to fuse multi-scale disparity and stereo network with different structures. SEDNet \cite{learningdistribution} proposes a soft-histogramming technique to align the distribution of prediction error and prediction uncertainty. Other works focus on improving the domain generalization performance \cite{masked,hvt,fcstereo,graftnet,chuah2022itsa} and domain adaptation performance \cite{adastereo,continual,stereogan,domaininvariant} of stereo matching networks.

Recognizing the need for increased efficiency and the integration of richer semantic information, multi-scale cost volume is used in stereo matching networks. HSMNet \cite{hsmnet} employs a pyramid of volumes to enable high-resolution stereo matching. AANet \cite{aanet} takes a lightweight approach, incorporating a feature pyramid and interactions with multi-scale correlation volumes. In a recent development, PCWNet \cite{pcwnet} introduces a volume fusion module that directly combines multi-scale 4D volumes. This innovation calculates a multi-level loss, ultimately expediting the model's convergence speed. 

\subsection{Geometric Structure Guided Stereo Matching}

Geometric structures have been used to guide stereo matching networks in previous works. EdgeStereo \cite{edgestereo} introduced an edge branch, integrating edge features into the stereo matching branch. Yang et al. \cite{yang2022edge} also advocated the use of edges for supervising stereo matching through multi-task learning. Additionally, surface normals have been another employed geometric structure to guide stereo matching. Zhang et al. \cite{zhang2017normal} suggested a joint prediction of a surface normal map and a raw disparity map, iteratively refining the disparity map with guidance from the surface normal map. Kusupati et al. \cite{kusupati2020normal} incorporated an extra surface normal branch to predict raw surface normals and employed an independently trained consistency module for refinement of both surface normals and raw disparities.

However, these approaches have two limitations. Firstly, they heavily depend on additional branches during inference to improve their performance, which impacts efficiency. In contrast, our proposed framework leverages geometric structures as constraints during training without incurring any overhead in inference, and shows improved performance. Secondly, the previous methods only made use of intra-view geometric structures, whereas our framework takes into account both intra-view and cross-view geometric knowledge through both geometric structure and their correspondences.

\subsection{Local Feature Matching}
Local feature matching aims at matching images that depict the same scene or object. This is usually done in a two-stage manner: first, detect interest points each associated with a visual descriptor from both images, and then match the detected interest points. 

Traditional interest points detection and description algorithms \cite{lowe2004distinctive,tradition1,tradition2,tradition3} rely on hand-crafted features. Recent research turns to Convolutional Neural Networks (CNN) for both interest point detection and description \cite{tyszkiewicz2020disk,liu2019gift,newpoint1,superretina}. SuperPoint \cite{superpoint} is a representative work of CNN-based interest point detector and descriptor, trained in a self-supervised manner. It constructs a synthetic shape dataset by rendering simplified 2D geometries, such as triangles, quadrilaterals, lines, and ellipses. The junctions of lines are labeled as interest points. These labels are then transferred to substantial real-world image datasets using a MagicPoint model and homographic adaptation. The self-supervised training process equips SuperPoint with the ability to be trained on various domains, enabling it to encode geometric information across diverse domains.

Traditional algorithms employ nearest-neighbor classifiers for matching interest points. Lowe's test \cite{lowe2004distinctive}, inlier classifiers \cite{inlineclassifier1,inlineclassifier2}, or geometric model fitting \cite{geomodel1,geomodel2} techniques are subsequently used to filter out unmatched points and incorrect correspondences. However, these methods are highly domain-specific and struggle to handle particularly challenging conditions. In contrast, deep learning-based matchers are trained to simultaneously match interest points and reject outliers. A notable example of this approach is SuperGlue \cite{sarlin2020superglue}, which leverages Transformers \cite{transformer} to conduct self-attention among interest points within the same view and cross-attention between interest points in different views. The final matching step involves solving an optimal transport problem. This approach allows SuperGlue to acquire robust insights into scene geometry and camera motion, rendering it resilient to extreme changes and effective in generalizing across various data domains. Subsequent works \cite{matcher1,matcher2,lindenberger2023lightglue} focused on making it more efficient and digging into its design details while keeping its strong matching capability. Consequently, these matchers are great sources of cross-view geometric knowledge. We set up constraints to let the stereo matching network learn geometric knowledge from a pre-trained interest point detector and a pre-trained interest point matcher.
\section{Method}

\begin{figure*}
    \centering
    \includegraphics[width=0.95\linewidth]{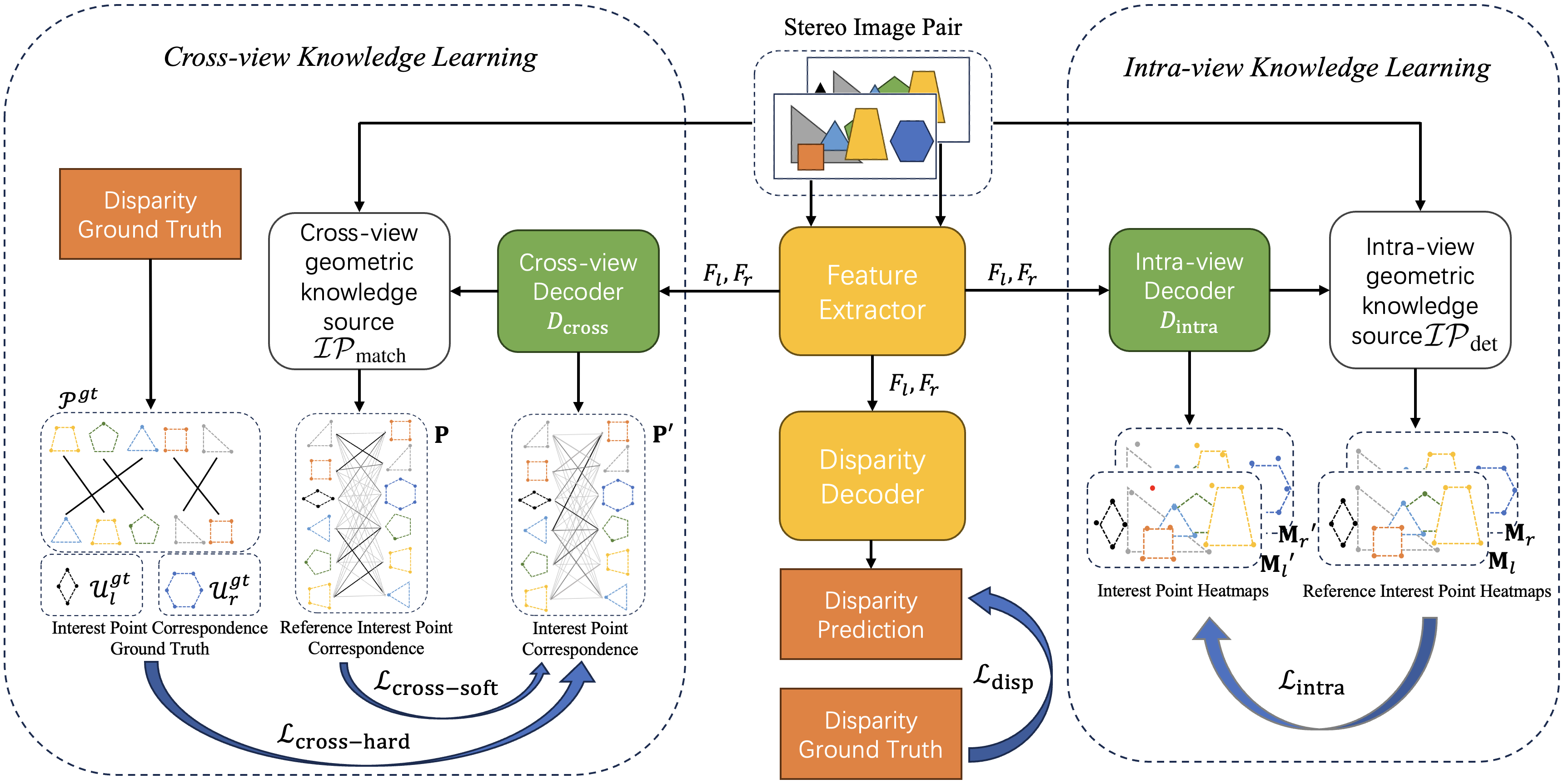}
    \caption{Overall structure of our proposed framework. The model architecture comprises three parts: stereo matching network, cross-view knowledge learning network, and intra-view knowledge learning network. The cross-view knowledge learning network introduces cross-view geometric knowledge by aligning the interest point correspondences $\mathbf{P}'$, $\mathbf{P}$ and $\mathcal{P}^{gt}$ using $\mathcal{L}_{\text{cross-hard}}$ and $\mathcal{L}_{\text{cross-soft}}$. The intra-view knowledge learning network introduces intra-view geometric knowledge through aligning the interest point maps $\mathbf{M}'$ and $\mathbf{M}$ using $\mathcal{L}_{\text{intra}}$. Note that the dotted lines are used just for clear illustration and are not used in our work.} 
    \label{fig:structure}
\end{figure*}

 We propose ICGNet to introduce geometric knowledge into stereo matching networks through geometric constraints. Intra-view geometric knowledge refers to the understanding that uncovers the geometric structures contained within a single image. Cross-view geometric knowledge refers to the comprehension that enables the alignment of geometric structures from one image to another. Models capable of uncovering and matching geometric structures harness the potential to convey intra-view and cross-view knowledge. Therefore, as depicted in \cref{fig:structure}, our framework introduces intra-view and cross-view geometric knowledge into the stereo matching network by aligning our network with intra-view and cross-view geometric knowledge sources. Although there are multiple choices (e.g., edges, points) to extract intra-view knowledge, we choose interest points as it is easy to further compute cross-view knowledge from these points using well-established methods \cite{sarlin2020superglue} while it is more challenging to match the edges for cross-view knowledge computation.  
We denote the left-view stereo image as $I_l$ and right-view stereo image as $I_r$, and the extracted left feature and right feature as $F_l$ and $F_r$.

\subsection{Intra-view Geometric Knowledge}

\textbf{Intra-view Geometric Knowledge Decoder} is to introduce intra-view geometric knowledge into the stereo matching network. We propose to leverage a pre-trained interest point detector $\mathcal{IP}_{\text{det}}$ \cite{superpoint} in our method. It takes the left-view image $I_l$ and the right-view image $I_r$ as input and generates interest maps $\mathbf{M}_l$ and $\mathbf{M}_r$, where each pixel is either labeled as interest point or non-interest point. These maps convey the intra-view geometric characteristics of the two images. We expect that the backbone features $F_l$ and $F_r$ preserve and unveil the intra-view geometric details present in both the left and right view images, akin to what the pre-trained interest point detector accomplishes. Consequently, we aim to extract two interest point maps, $\mathbf{M}'_l$ and $\mathbf{M}'_r$, from $F_l$ and $F_r$ using a trainable intra-view decoder $D_{\text{intra}}$, aligning with $\mathbf{M}_l$ and $\mathbf{M}_r$ respectively. For the design of $D_{\text{intra}}$, we use a stack of two bottleneck convolutional blocks. Note that decoder $D_{\text{intra}}$ is discarded during the test stage.

\textbf{Intra-view geometric loss} $\mathcal{L}_{\text{intra}}$ is computed to enforce this alignment:

\begin{equation}
    \mathcal{L}_{\text{intra}} = \frac{1}{2}\mathcal{L}(\mathbf{M}'_l, \mathbf{M}_l) + \frac{1}{2}\mathcal{L}(\mathbf{M}_r', \mathbf{M}_r),
\end{equation}
where we compute a focal loss $\mathcal{L}$ due to unbalanced positive and negative samples of interest points.

\subsection{Cross-view Geometric Knowledge}

\textbf{Cross-view Geometric Knowledge Decoder} enables the stereo matching network to acquire cross-view geometric knowledge by framing it as a learning task. This task is to align the stereo matching network with two complementary knowledge sources: a pre-trained interest point matcher denoted as $\mathcal{IP}_{\text{match}}$ where we use \cite{sarlin2020superglue} in our method, and the ground truth matching between interest points. 

The pre-trained matcher processes two sets of interest points extracted from stereo image pair $I_l$ and $I_r$, denoted as $p_l$ and $p_r$, each containing $m$ and $n$ points respectively, along with their associated descriptors. It outputs an assignment matrix $\mathbf{P} \in [0, 1]^{(m+1)\times (n+1)}$. This matrix includes an extra row and column to account for the possibility that some points do not match any counterpart in the other view. The assignment matrix $\mathbf{P}$ elucidates the cross-view geometric knowledge via the relations between the matching probabilities of the two sets of points. 

Furthermore, the ground truth disparities can serve as another source of knowledge since it provide accurate matching between these points. Ground truth matching pairs between the points $\mathcal{P}^{gt}=\{(i, j)\} \subset p_l \times p_r$ ($\times$ refers to Cartesian product) and unmatched points in left image $\mathcal{U}_l^{gt} \subseteq p_l$ and right image $\mathcal{U}_r^{gt} \subseteq p_r$ are formed by warping interest points from the left image to the right image using the disparity. 

The ground truth is more precise compared to the assignment matrix provided by the interest point matcher. In the ground truth, the relationships between points are binary, classified simply as either a match or a mismatch. On the other hand, the assignment matrix presents matching scores as continuous values ranging from 0 to 1, implicitly indicating the degree of similarity between the points. This makes the two approaches complementary: one provides precise matching categorization, which is `harder', while the other offers a fine-grained assessment of similarity, which is `softer'.

To align the stereo matching network with these two sources of cross-view geometric knowledge, a trainable cross-view decoder denoted as $D_{\text{cross}}$ is proposed. $D_{\text{cross}}$ first use as an Multilayer Perceptron (MLP) encoder \cite{gehring2017convolutional,transformer} followed by four alternative self-attention and cross-attention layers \cite{transformer}. The self attention layers operate on points within the same image, while the cross-attention layer operate on points within different images. The decoder is responsible for decoding an assignment matrix $\mathbf{P}'$. This decoded matrix is trained to closely match its corresponding reference assignment matrix $\mathbf{P}$ created by a pre-trained interest point matcher $\mathcal{IP}_{\text{match}}$ and the ground truth matching pairs $\mathcal{P}^{gt}$ and unmatched points $\mathcal{U}_l^{gt}$, $\mathcal{U}_r^{gt}$ from left and right images respectively, serving as the supervision pseudo ground truth during training. The points input $p_{l}$ and $p_{r}$ to the decoder $D_{\text{cross}}$ are identical to the input to $\mathcal{IP}_{\text{match}}$, but with different descriptors bi-linearly sampled from features $F_l$ and $F_r$ extracted by stereo backbone such that the gradient can flow back to the stereo network. Note that decoder $D_{\text{cross}}$ is discarded during the test stage.

\textbf{Soft cross-view geometric loss} $\mathcal{L}_{\text{cross-soft}}$ is proposed to introduce the cross-view geometric knowledge from the reference assignment matrix $\mathbf{P}$ into the predicted assignment matrix $\mathbf{P}'$.  

\begin{equation}
    \begin{split}
        \mathcal{L}_{\text{cross-soft}}
        &=\frac{1}{m} \sum_{i=1}^m \mathcal{KL}\left(\mathbf{P}_{i, \cdot}/||\mathbf{P}_{i, \cdot}||_1, \mathbf{P}'_{i, \cdot}/||\mathbf{P}'_{i, \cdot}||_1\right)\\
        &+\frac{1}{n} \sum_{j=1}^n \mathcal{KL}\left(\mathbf{P}_{\cdot, j}/||\mathbf{P}_{\cdot, j}||_1,\mathbf{P}'_{\cdot, j}/||\mathbf{P}'_{\cdot, j}||_1\right),
    \end{split}
\end{equation}
where $\mathcal{KL}$ stands for KL-divergence, and $\mathbf{P}_{i, \cdot}$ stands for the $i^{th}$ row of $\mathbf{P}$, $\mathbf{P}_{\cdot, j}$ stands for the $j^{th}$ column of $\mathbf{P}$. In this work, we use the interest points filtered from interest point maps $\mathbf{M}_l$, $\mathbf{M}_r$ extracted by pre-trained interest point detector \cite{superpoint} by non maximum suppression   as $p_l$ and $p_r$.

\textbf{Hard cross-view geometric loss} is further computed based on the ground truth matching between interest points $\mathcal{P}^{gt}$, unmatched points in the left image $\mathcal{U}_l^{gt}$, and unmatched points in the right image $\mathcal{U}_r^{gt}$. Hard cross-view geometric loss is utilized as a negative log-likelihood loss balanced between matched points and unmatched points to align $\mathbf{P}'$ and $\mathcal{P}^{gt}$, $\mathcal{U}_l^{gt}$, $\mathcal{U}_r^{gt}$:

\begin{equation}
    \begin{split}
        \mathcal{L}_{\text{cross-hard}} = 
        &-\frac{1}{|\mathcal{P}^{gt}|}\sum_{(i, j)\in \mathcal{P}^{gt}}\log \mathbf{P}'_{i,j}\\ 
        &- \frac{1}{|\mathcal{U}_l^{gt}|}\sum_{i\in \mathcal{U}_l}\log \mathbf{P}'_{i,n+1}\\
        &-\frac{1}{|\mathcal{U}_r^{gt}|}\sum_{j\in \mathcal{U}_r}\log \mathbf{P}'_{m+1,j}.
    \end{split}
\end{equation}

\subsection{Loss}

We integrate our proposed $\mathcal{L}_{\text{intra}}$, $\mathcal{L}_{\text{cross-soft}}$, and $\mathcal{L}_{\text{cross-hard}}$ with disparity loss $L_{\text{disp}}$, enabling simultaneously estimating disparity and learning intra-view and cross-view geometric structure knowledge from local feature matching models. The disparity loss, $L_{\text{disp}}$, is determined by the loss function utilized by the base models. The total loss is defined as:
\begin{equation}
    \begin{split}
        \mathcal{L}_{\text{total}} &= \mathcal{L}_{\text{disp}} + \lambda_{\text{intra}}\cdot\mathcal{L}_{\text{intra}} + \lambda_{\text{cross-soft}}\cdot\mathcal{L}_{\text{cross-soft}} \\
        &+ \lambda_{\text{cross-hard}}\cdot\mathcal{L}_{\text{cross-hard}},
    \end{split}
\end{equation}
where $\mathcal{L}_{\text{intra}}$, $\mathcal{L}_{\text{cross-soft}}$, $\mathcal{L}_{\text{cross-hard}}$ are weighting terms. The loss weights are empirically set as $\mathcal{L}_{\text{intra}}=100$, $\mathcal{L}_{\text{cross-soft}}=0.5$, $\mathcal{L}_{\text{cross-hard}}=0.5$ such that they are comparable and none of the items would dominate the results.
\section{Experiments}

\begin{table*}[tb]\small
    \centering
    \begin{tabular}{cccccccc}
    \toprule
        Method   & PSMNet \cite{psmnet} & GANet \cite{ganet} & ACVNet \cite{acvnet} & DLNR \cite{dlnr} & GwcNet \cite{gwcnet} & IGEV-Stereo \cite{igevstereo} & Ours\\ 
        \hline
        EPE(px)$\downarrow$ & 1.09 & 0.84 & 0.48 & 0.48 & 0.765 & 0.479 & \textbf{0.447}\\
    \bottomrule
    \end{tabular}
    \caption{Comparison with state-of-the-art models on the SceneFlow dataset. Our method outperforms most of the state-of-the-art methods.}
    \label{tab:sota}
\end{table*}

\begin{table*}[tb]\small
    \centering
    \begin{tabular}{l|cccc|ccc|ccc}
        \toprule
            \multirow{2}{*}{Method} & \multicolumn{4}{c|}{KITTI 2012} & \multicolumn{3}{c|}{KITTI 2015 (all pixels)} & \multicolumn{3}{c}{KITTI 2015 (noc pixels)} \\
        \cline{2-11}
            &2-noc$\downarrow$ &2-all$\downarrow$ &3-noc$\downarrow$ &3-all$\downarrow$ &D1-bg$\downarrow$ &D1-fg$\downarrow$ &D1-all$\downarrow$ &D1-bg$\downarrow$ &D1-fg$\downarrow$ &D1-all$\downarrow$ \\
        \hline
        \hline
            EdgeStereo \cite{edgestereo} & 2.32 & 2.88 & 1.46 & 1.83 & 1.84 & 3.30 & 2.08 & 1.69 & 2.94 & 1.89\\
            LEAStereo \cite{leastereo} & 1.90 & 2.39 & 1.13 & 1.45 & 1.40 & 2.91 & 1.65 & 1.29 & 2.65 & 1.51\\
            ACVNet \cite{acvnet} & 1.83 & 2.35 & 1.13 & 1.47 & \textbf{1.37} & 3.07 & 1.65 & 1.26 & 2.84 & 1.52\\
            CREStereo \cite{crestereo} & 1.72 & 2.18 & 1.14 & 1.46 & 1.45 & 2.86 & 1.69 & 1.33 & 2.60 & 1.54\\
            RAFT-Stereo \cite{raftstereo} & 1.92 & 2.42 & 1.30 & 1.66 & 1.58 & 3.05 & 1.82 & 1.45 & 2.94 & 1.69 \\
            AcfNet \cite{acfnet} & 1.83 & 2.35 & 1.17 & 1.54 & 1.51 & 3.80 & 1.89 & 1.36 & 3.49 & 1.72\\
            HITNet \cite{tankovich2021hitnet} & 2.00 & 2.65 & 1.41 & 1.89 & 1.74 & 3.20 & 1.98 & 1.54 & 2.72 & 1.74\\
            GANet \cite{ganet} & 1.89 & 2.50 & 1.19 & 1.60 & 1.48 & 3.46 & 1.81 & 1.34 & 3.11 & 1.63\\
            GwcNet \cite{gwcnet}           & 2.16 & 2.71 & 1.32 & 1.70 & 1.74 & 3.93 & 2.11 & 1.61 & 3.49 & 1.92\\
            IGEV-Stereo \cite{igevstereo}      & 1.71 & 2.17 & 1.12 & 1.44 & 1.38 & 2.67 & 1.59  & 1.27 & 2.62 & 1.49\\  \hline
            Ours   & \textbf{1.70}  & \textbf{2.14} & \textbf{1.10} & \textbf{1.41} & 1.38 & \textbf{2.55} & \textbf{1.57} & \textbf{1.26} & \textbf{2.56} & \textbf{1.47} \\
        \bottomrule
    \end{tabular}
    \caption{Finetuning results on KITTI 2012 and KITTI 2015 benchmarks. Our ICGNet achieves state-of-the-art performance on both benchmarks and ranks \textbf{$1^{st}$} on the KITTI 2015 stereo matching benchmark at the time of submission.}
    \label{tab:sota_kitti}
\end{table*}

\begin{table*}[tb]\small
    \centering
    \begin{tabular}{l|cc|cc|cc}
        \toprule
        \multirow{2}{*}{Method} & \multicolumn{2}{c|}{KITTI 2012} &  \multicolumn{2}{c|}{KITTI 2015} & \multicolumn{2}{c}{Middlebury 2014(H)} \\
        \cline{2-7}
        & EPE(px)$\downarrow$ & $>$3px(\%)$\downarrow$ & EPE(px)$\downarrow$ & $>$3px(\%)$\downarrow$ & EPE(px)$\downarrow$ & $>$2px(\%)$\downarrow$\\ 
        \hline
        \hline
            PSMNet \cite{psmnet} & 2.69 & 15.1 & 3.17 & 16.3 & 7.65 & 34.2\\
            CFNet \cite{cfnet}  & 1.04 & 5.2 & 1.71 & 6.0 & 3.24 & 15.4\\
            GANet \cite{ganet} & 1.93 & 10.1 & 2.31 & 11.7 & 5.41 & 20.3\\
            DSMNet \cite{domaininvariant} & 1.26 & 6.2 & 1.46 & 6.5 & 2.62 & 13.8\\
            IGEV-Stereo \cite{igevstereo}     & 1.04 & 5.18 & 1.21 & 6.03 & 0.91 & 7.27\\ 
        \hline
            Ours  & \textbf{0.99} & \textbf{4.87} & \textbf{1.16} & \textbf{5.96} & \textbf{0.82} & \textbf{6.73}\\
        \bottomrule
    \end{tabular}
    \caption{Cross-domain generalization from SceneFlow to real-world datasets.}
    \label{tab:cross_generalization}
\end{table*}

\begin{table}[tb]\scriptsize
  \centering
  \begin{tabular}{l|ccc|cc}
    \toprule
    \multirow{2}{*}{Baseline} & \multicolumn{3}{c|}{Components} & \multicolumn{2}{c}{SceneFlow} \\
    \cline{2-6}
    & $\mathcal{L}_{\text{intra}}$ & $\mathcal{L}_{\text{cross-soft}}$ & $\mathcal{L}_{\text{cross-hard}}$ & EPE(px)$\downarrow$ & $>$3px(\%)$\downarrow$ \\
    \hline 
    \hline
    \multirow{4}{*}{GwcNet} &  &  &  & 0.765 & 3.30 \\
     & \checkmark &  &  & 0.615 & 2.61 \\
     & \checkmark & \checkmark &  & 0.608 & 2.56 \\
     & \checkmark & \checkmark & \checkmark & \textbf{0.597} & \textbf{2.55} \\
    \hline
    \multirow{4}{*}{IGEV-Stereo} &  &  &  & 0.479 & 2.48 \\
     & \checkmark &  &  & 0.452 & 2.37 \\
     & \checkmark & \checkmark &  & 0.449 & 2.36 \\
     & \checkmark & \checkmark & \checkmark & \textbf{0.447} & \textbf{2.34} \\
    \bottomrule
  \end{tabular}
  \caption{Ablation studies of proposed components on SceneFlow dataset on two baseline approaches GwcNet and IGEV-Stereo.}
  \label{tab:ablation}
\end{table}

\begin{table*}[tb]\small
    \centering
    \begin{tabular}{l|c|cccc|ccc|ccc}
        \toprule
            \multirow{2}{*}{Baseline} & \multirow{2}{*}{ICGNet} & \multicolumn{4}{c|}{KITTI 2012} & \multicolumn{3}{c|}{KITTI 2015 (all pixels)} & \multicolumn{3}{c}{KITTI 2015 (noc pixels)} \\
        \cline{3-12}
            & &2-noc$\downarrow$ &2-all$\downarrow$ &3-noc$\downarrow$ &3-all$\downarrow$ &D1-bg$\downarrow$ &D1-fg$\downarrow$ &D1-all$\downarrow$ &D1-bg$\downarrow$ &D1-fg$\downarrow$ &D1-all$\downarrow$ \\
        \hline
        \hline
            \multirow{2}{*}{GwcNet \cite{gwcnet}}  &  & 2.16 & 2.71 & 1.32 & 1.70 & 1.74 & 3.93 & 2.11 & 1.61 & 3.49 & 1.92\\
             & \checkmark & \textbf{1.98} & \textbf{2.54} & \textbf{1.25} & \textbf{1.63} & \textbf{1.62} & \textbf{3.90} & \textbf{2.00} & \textbf{1.50} & \textbf{3.56} & \textbf{1.84}\\
        \hline
            \multirow{2}{*}{IGEV-Stereo \cite{igevstereo}}   &   & 1.71 & 2.17 & 1.12 & 1.44 & 1.38 & 2.67 & 1.59 & 1.27 & 2.62 & 1.49\\ 
            &  \checkmark & \textbf{1.70}  & \textbf{2.14} & \textbf{1.10} & \textbf{1.41} & 1.38 & \textbf{2.55} & \textbf{1.57} & \textbf{1.26} & \textbf{2.56} & \textbf{1.47} \\
        \bottomrule
    \end{tabular}
    \caption{Ablation studies on KITTI 2012 and KITTI 2015 benchmarks.}
    \label{tab:ablation_kitti}
\end{table*}

\begin{table}[t]\small
  \centering
  \begin{tabular}{cc|ccc}
    \toprule
    Intra-view  & Cross-view   & \multicolumn{3}{c}{SceneFlow EPE} \\
    \cline{3-5}
    $\mathcal{L}_{\text{intra}}$ & $\mathcal{L}_{\text{cross-soft}} + \mathcal{L}_{\text{cross-hard}}$ & noc(px)$\downarrow$ & occ(px)$\downarrow$ & $\frac{\Delta\text{noc}}{\Delta\text{occ}}$\\
    \hline 
    \hline
    &  & 0.413 & 2.58 & / \\
    \checkmark &  & 0.334 & 2.08 & 6.33 \\
    \checkmark & \checkmark & 0.318 & 2.02 & 3.75 \\
    \bottomrule
  \end{tabular}
  \caption{Ablation studies on occluded and non-occluded areas on the SceneFlow dataset using GwcNet baseline model. Our method improves disparity prediction accuracy in both occluded areas and non-occluded areas. 
  Compared with cross-view geometric knowledge, intra-view geometric knowledge improves more in occluded regions, consistent with the intuition that occluded areas rely on their surrounding geometric relations to infer their disparities.}
  \label{tab:intra-cross}
\end{table}

\begin{figure*}[tb]
    \centering
    \includegraphics[width=0.95\textwidth]{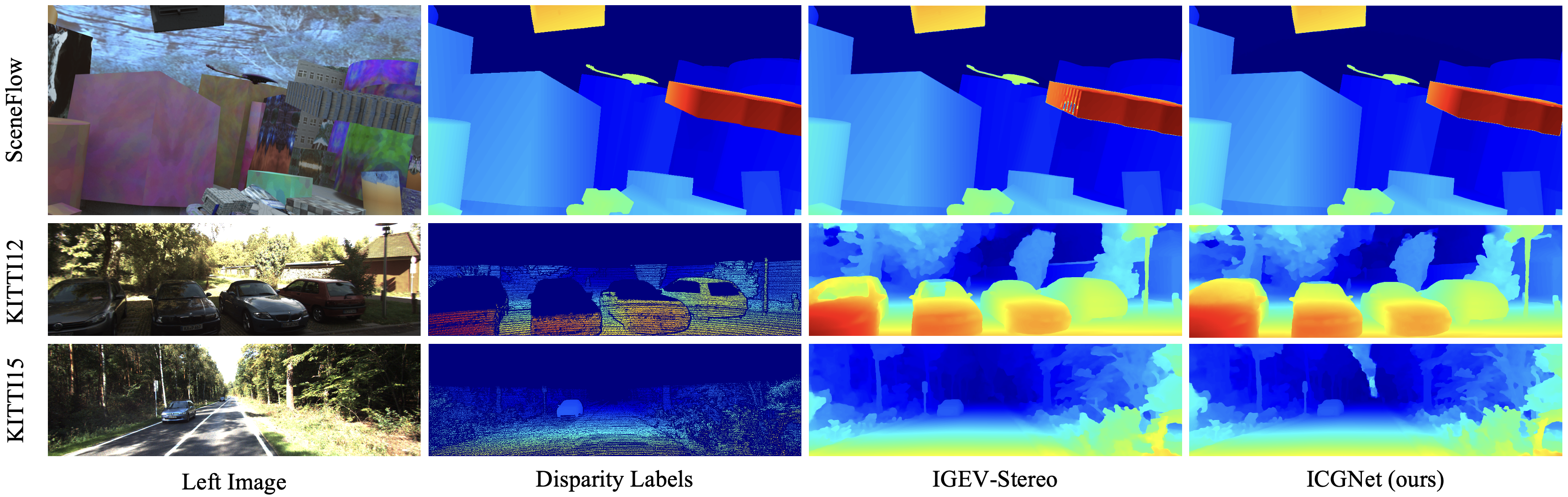}
    \caption{Qualitative results of ICGNet (ours) compared with the state-of-the-art method IGEV-Stereo.}
    \label{fig:sota_qualitative}
\end{figure*}


\begin{table}[htb]\small
    \centering
    \begin{tabular}{c|c|cc}
        \toprule
        \multirow{2}{*}{Decoder} & \multirow{2}{*}{Layers} & \multicolumn{2}{c}{SceneFlow} \\
        \cline{3-4}
        & & EPE(px)$\downarrow$ & $>$3px$\downarrow$ \\
        \hline
        \hline
        \multirow{3}{*}{intra-view} & 1 & 0.631 & 2.74  \\
         & 2 & \textbf{0.615} & \textbf{2.61} \\
         & 4 & 0.623 & 2.68 \\
        \hline
        \multirow{3}{*}{cross-view} & 0 & 0.617 & 2.61 \\
         & 4 & \textbf{0.597} & \textbf{2.55} \\
         & 6 & 0.606 & 2.58 \\
        \bottomrule
    \end{tabular}
    \caption{Comparison of performance of different decoder complexities.}
    \label{tab:decoder}
\end{table}


\textbf{SceneFlow} dataset \cite{sceneflow} constitutes a synthetic dataset featuring diverse random objects. It contains 35,454 training pairs and 4,370 test pairs with dense disparity labels. Finalpass of the SceneFlow dataset that contains motion blur and defocuses is used.

\textbf{KITTI 2012 \cite{kitti12} \& KITTI 2015 \cite{kitti15}} datasets consist of stereo image pairs of real-world driving scenarios. KITTI 2012 contains 194 training and 195 testing image pairs. KITTI 2015 comprises 200 training and 200 testing image pairs. The stereo pairs in both KITTI 2012 and KITTI 2015 possess resolutions of 1226×370 and 1242×375, respectively. Notably, both datasets furnish sparse disparity maps as part of their annotations.

\textbf{Middlebury 2014} \cite{middlebury} is an stereo dataset consisted of indoor stereo images. It comprises 15 training and 15 testing pairs. All the images are available in three different resolutions. We use the Middlebury 2014 dataset with half resolution to test cross-domain generalization performance. 

\textbf{Evaluation metrics.} We use end point error (EPE), the percentage of errors larger than 3px ($>$3px) on the SceneFlow dataset, and use EPE and D1-3px on KITTI datasets as evaluation metrics for disparity prediction. D1-3px refers to the percentage of disparity estimations with both absolute larger than 3px and relative error greater than 5\%. We use EPE and the percentage of errors larger than 2px ($>$2px) on Middlebury 2014 dataset. 

\subsection{Implementation Details}
Our ICGNet is inherently compatible with most stereo matching models since we do not change the structure of the network during inference. Specifically, our framework’s compatibility expands to encompass all stereo matching models incorporating a common feature extractor and a disparity decoder. We experiment with our ICGNet on different baseline models GwcNet \cite{gwcnet} and IGEV-Stereo \cite{igevstereo}. We train the models end-to-end for all experiments. We pre-train the models on the SceneFlow training set. For the evaluation of cross-domain generalization, we directly test the sceneflow pre-trained models' performance on the KITTI 2012 training set, KITTI 2015 training set, and Middlebury 2014 training set. For evaluation of KITTI 2012 and KITTI 2015 benchmarks, we finetune SceneFlow pre-trained model on the mix of KITTI 2012 and KITTI 2015 training sets. For GwcNet, we use the GwcNet-g setting for KITTI 2015 pre-training and the GwcNet-gc setting for other experiments, consistent with the original work for a fair comparison. We train the model for 64 epochs on SceneFlow and 1000 epochs on KITTI. The initial learning rate is set to 0.001. The learning rate is decayed by a factor of 2 after epochs 20, 32, 40, 48, and 56 on SceneFlow, and decayed by a factor of 2 after epochs 200, 400, 600, 700, and 800 on KITTI. For IGEV-Stereo, we train the model for 80 epochs on SceneFlow and 1000 epochs on KITTI. The initial learning rate is set as 0.0002 and a one-cycle learning rate schedule is used. We maintain all other settings consistent with the original works for both models. All the codes in this paper would be released in Github.

\begin{figure*}[tb]
    \centering
    \includegraphics[width=0.95\textwidth]{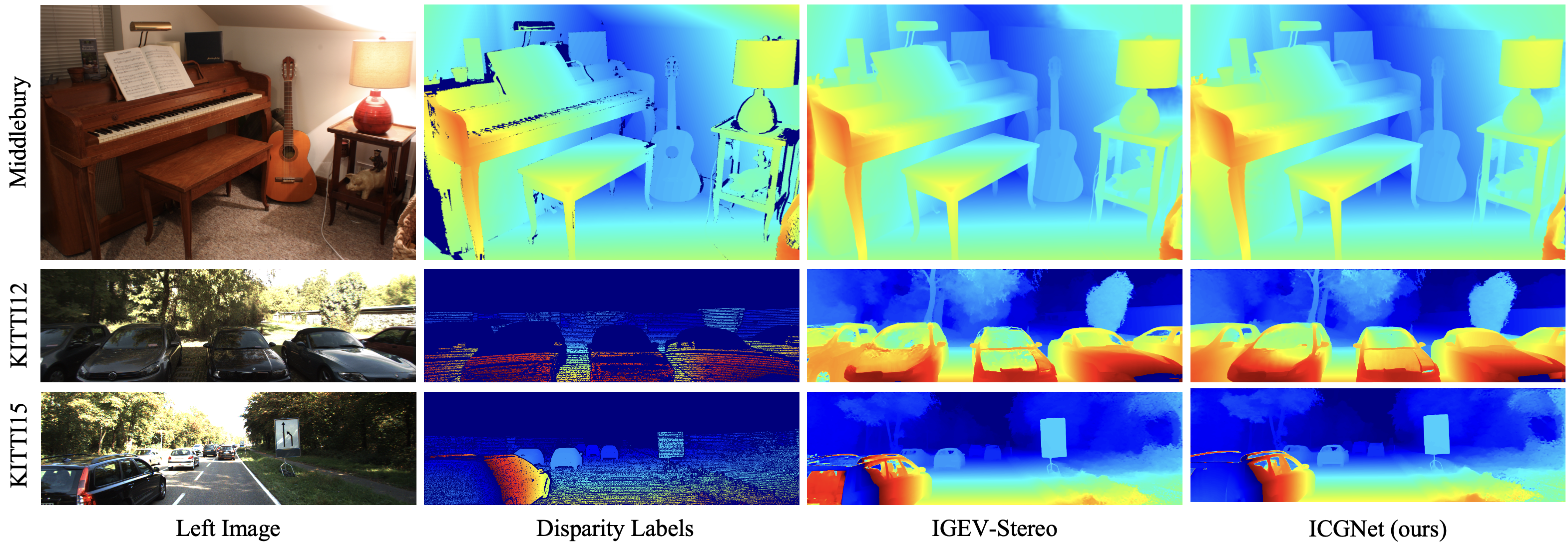}
    \caption{Qualitative results of cross-domain generalization of ICGNet (ours) compared with state-of-the-art baseline IGEV-Stereo \cite{igevstereo}.}
    \label{fig:cross_qualitative}
\end{figure*}

\subsection{Comparison with State-of-the-art}

To evaluate the effectiveness of our proposed method, we use the latest IGEV-Stereo \cite{igevstereo} as the base model and integrate it with ICGNet to compare with state-of-the-art stereo mathing methods. \cref{tab:sota} presents a comparison of our method with state-of-the-art on SceneFlow. Our proposed method achieved state-of-the-art performance on the SceneFlow dataset. We further finetune the SceneFlow pre-trained model on KITTI12 \cite{kitti12} and KITTI15 \cite{kitti15} datasets, and compare the performance of ICGNet with state-of-the-art methods on KITTI 2012 and KITTI 2015 benchmarks. Results are shown in \cref{tab:sota_kitti}. Our ICGNet outperforms state-of-the-art methods in most scenarios.

Besides quantitative results, qualitative experiments are conducted to show the effectiveness of our proposed method. \cref{fig:sota_qualitative} visualizes the disparity predictions of our methods. Compared to the state-of-the-art method IGEV-Stereo, our method performs better in reflective regions and unlabelled backgrounds.

\subsection{Cross-domain Generalization}

We carry out experiments to demonstrate that our approach enhances the domain-generalization capacity of base models, transitioning from the synthetic SceneFlow dataset to real-world datasets. As illustrated in \cref{tab:cross_generalization}, our proposed ICGNet method boosts the cross-domain generalization performance for IGEV-Stereo which is a strong baseline for cross-domain generalization. Our ICGNet outperforms IGEV-Stereo by 4.8\% on EPE in KITTI 2012 and 4.1\% in KITTI 2015. Furthermore, from the qualitative results in \cref{fig:cross_qualitative}, we can see that our method outperforms IGEV-Stereo in reflective regions and occluded regions.

\subsection{Ablation Studies}
We conduct ablation studies to evaluate the effectiveness of our proposed method. To show the generalization of the intra-view and cross-view knowledge learning, two base methods are used: GwcNet \cite{gwcnet} and IGEV-Stereo \cite{igevstereo}. 

\textbf{Effectiveness of Intra-view and Cross-view Knowledge.} To verify that our proposed components all contribute to improving the performance, we provide quantitative results on the SceneFlow dataset in \cref{tab:ablation}. The proposed ICGNet framework introduces an intra-view geometric loss $\mathcal{L}_{\text{intra}}$, a soft cross-view geometric loss $\mathcal{L}_{\text{cross-soft}}$ and a hard cross-view geometric loss $\mathcal{L}_{\text{cross-hard}}$. The intra-view geometric loss allows the model to learn intra-image geometric knowledge, and the soft cross-view geometric loss and hard cross-view geometric loss allow the model to learn cross-view geometric knowledge. Notably, our method demonstrates an improvement of over 20\% (0.597 vs 0.765) in EPE and over 20\% improvement in 3px-error. When applied to the strong base model IGEV-Stereo, our method still shows solid improvements over both EPE and 3px-error. The SceneFlow pre-trained full models of both base models are then finetuned on KITTI 2012 and KITTI 2015 datasets and results are shown in \cref{tab:ablation_kitti}. Our method consistently outperforms the performance of baseline models.

\textbf{Impact on Occluded and Non-occluded Regions.} We further conduct experiments on our method's impact on occluded pixels and non-occluded pixels separately. \cref{tab:intra-cross} shows that our method consistently improves disparity quality in both occluded regions and non-occluded regions. Furthermore, since occluded regions are areas without correspondence in the other figures, intuitively the prediction of disparities on such pixels relies more on intra-view geometric knowledge; on the contrary, predicting disparities at non-occluded pixels relies more on cross-view geometric knowledge \cite{igevstereo}. Our experiment results are consistent with this intuition, validating our formulation of the process of extracting interest points as intra-view geometric knowledge and the process of extracting their correspondences as cross-view geometric knowledge.  



\textbf{Complexity of intra-view decoder and cross-view decoder.} We conduct experiments to evaluate the impact of the complexity of the intra-view decoder $D_{\text{intra}}$ and cross-view decoder $D_{\text{cross}}$ on the models' performance. We keep GwcNet as the base model and change the number of layers of the decoders. The number of layers of the intra-view decoder refers to the number of stacked bottleneck convolutional blocks, and the number of layers of the cross-view decoder refers to the number of alternating attention layers. For the cross-view decoder with zero layers, we are directly using cosine similarities between the interest point features plus a softmax layer to form an assignment matrix $\mathbf{P}'$. Results on the SceneFlow dataset with GwcNet \cite{gwcnet} as baseline are shown in \cref{tab:decoder}. Results have shown that an intermediate number of decoder layers for both intra-view and cross-view decoder yields the best results. This may be due to a too-complex decoder will take up too much of the task of learning the geometric knowledge, making the feature backbone not learn sufficient knowledge, while a too-simple decoder is not sufficient to map the features into the geometric structures, leaving too much for the feature backbone and make it sub-optimal in disparity prediction task.

\section{Conclusion}

In this work, we introduce ICGNet, a novel approach designed to infuse stereo matching networks with both intra-view and cross-view geometric knowledge. Utilizing interest points and their corresponding matches derived from local feature matching models, we provide rich sources of geometric knowledge to educate the stereo matching network. Our extensive experimental evaluations demonstrate that this methodology enhances the performance of state-of-the-art stereo matching models, boosting both their accuracy and their ability to generalize across diverse domains. 

\section*{Acknowledgement} 
This work was supported by the Agency for Science, Technology and Research (A*STAR) under its MTC Programmatic
 Funds (Grant No. M23L7b0021).
{
    \small
    \bibliographystyle{ieeenat_fullname}
    \bibliography{main}
}


\end{document}